# A Probabilistic Calculus of Actions

**Judea Pearl**
Cognitive Systems Laboratory
Computer Science Department
University of California, Los Angeles, CA 90024
*judea@cs.ucla.edu*

## Abstract

We present a symbolic machinery that admits both probabilistic and causal information about a given domain and produces probabilistic statements about the effect of actions and the impact of observations. The calculus admits two types of conditioning operators: ordinary Bayes conditioning, $P(y|X = x)$, which represents the observation $X = x$, and causal conditioning, $P(y|do(X = x))$, read the probability of $Y = y$ conditioned on holding $X$ constant (at $x$) by deliberate action. Given a mixture of such observational and causal sentences, together with the topology of the causal graph, the calculus derives new conditional probabilities of both types, thus enabling one to quantify the effects of actions (and policies) from partially specified knowledge bases, such as Bayesian networks in which some conditional probabilities may not be available.

## 1 INTRODUCTION

Probabilistic methods, especially those based on graphical models, have proven useful in tasks of prediction, abduction and belief revision [Pearl 1988, Heckerman 1990, Goldszmidt 1992, Darwiche 1993]. In planning, however, they are less popular,[1] partly due to the unsettled, strange relationship between probability and actions. In principle, actions are not part of standard probability theory, and understandably so: probabilities capture normal relationships in the world, while actions represent interventions that perturb those relationships. It is no wonder, then, that actions are treated as foreign entities throughout the literature on probability and statistics; they serve neither as arguments of probability expressions nor as events for conditioning such expressions. Even in the decision theoretic literature, where actions are the target of optimization, the symbols given to actions serve merely as indices for distinguishing one probability function from another, not as propositions that specify the immediate effects of the actions. As a result, if we are given two probabilities, $P_A$ and $P_B$, denoting the probabilities prevailing under actions $A$ or $B$, respectively, there is no way we can deduce from this input the probability $P_{A \wedge B}$ corresponding to the joint action $A \wedge B$, or any Boolean combination of the propositions $A$ and $B$. This means that, in principle, the impact of all anticipated joint actions would need to be specified in advance—an insurmountable task by any standard.

The peculiar status of actions in probability theory can be seen most clearly in comparison to the status of observations. By specifying a probability function $P(s)$ on the possible states of the world, we automatically specify how probabilities would change with every conceivable observation $e$, since $P(s)$ permits us to compute (using Bayes rule) the posterior probabilities $P(E|e)$ for every pair of events $E$ and $e$. However, specifying $P(s)$ tells us nothing about how our probabilities should be revised as a response to an external action $A$. In general, if an action $A$ is to be described as a function that takes $P(s)$ and transforms it to $P_A(s)$, then Bayesian conditioning is clearly inadequate for encoding this transformation. For example, consider the statements: "I have observed the barometer reading to be $x$" and "I intervened and set the barometer reading to $x$". If processed by Bayes conditioning on the event "the barometer reading is $x$", these two reports would have the same impact on our current probability function, yet we certainly do not consider the two reports equally informative about an incoming storm.

The engineering solution to this problem is to include the acting agents as variables in the analysis, construct a distribution function including the behavior of those agents, and infer the effect of the action by conditioning those "agent variables" to a particular mode of behavior. Thus, for example, the agent manipulating the barometer would enter the system as a variable such as, "Squeezing the barometer" or "Heating the barometer". After incorporating this variable into the probability distribution, we could infer the impact of manipulating the barometer by simply conditioning the distribution on the event "Squeezing the barometer reached level $x$". This is, in effect, the solution adopted in influence diagrams (IDs), the graphical tool proposed for decision theory [Howard & Matheson 1981, Shachter 1986]. Each anticipated action is represented as a variable (a node in the diagram), and its impact on other variables is assessed and encoded in terms of conditional probabilities, similar to the impact of any other parent node in the diagram.

---

[1] Works by Dean & Kanazawa [1989] and Kushmerick et al. [1993] notwithstanding.

The difficulties with this approach are twofold. First, the approach is procedural (rather than declarative) and therefore lacks the semantics necessary for supporting symbolic derivations of the effects of actions. We will see in Section 3 that such derivations become indispensable in processing partially specified diagrams. Second, the need to anticipate and represent all relevant actions in advance renders the elicitation process unduly cumbersome. In circuit diagnosis, for example, it would be awkward to represent every conceivable act of component replacement (similarly, every conceivable connection to a voltage source, current source, etc.) as a node in the diagram. Instead, the effects of such actions are implicit in the circuit diagram itself and can be inferred directly from the (causal) Bayesian network that represents the workings of the circuit.[2] We therefore concentrate our discussion on knowledge bases where actions are not represented explicitly. Rather, each action will be indexed by a proposition which describes the condition we wish to enforce directly. Indirect consequences of these conditions will be inferred from the causal relationships among the variables represented in the knowledge base.

As an alternative to Bayesian conditioning, philosophers [Lewis 1976] have studied another probability transformation called "imaging" which was deemed useful in the analysis of subjunctive conditionals and which more adequately represents the transformations associated with actions. Whereas Bayes conditioning $P(s|e)$ transfers the entire probability mass from states excluded by $e$ to the remaining states (in proportion to their current $P(s)$), imaging works differently: each excluded state $s$ transfers its mass individually to a select set of states $S^*(s)$, considered "closest" to $s$. While providing a more adequate and general framework for actions, imaging leaves the precise specification of the selection function $S^*(s)$ almost unconstrained. The task of formalizing and representing these specifications can be viewed as the probabilistic version of the infamous *frame problem* and its two satellites, the *ramification* and *concurrent actions* problems.

An assumption commonly found in the literature is that the effect of an elementary action $do(q)$ is merely to change $\neg q$ to $q$ where the current state satisfies $\neg q$ and, otherwise, to leave things unaltered.[3] We can call this assumption the "delta" rule, variants of which are embedded in STRIPS as well as in probabilistic planning systems. In BURIDAN [Kushmerick et al. 1993], for example, every action is specified as a probabilistic mixture of several elementary actions, each operating under the delta rule.

The problem with the delta rule and its variants is that they do not take into account the indirect ramifications of an action such as, for example, those triggered by chains of causally related events. To handle such ramifications, we must construct a causal theory of the domain, specifying which event chains are likely to be triggered by a given action (the ramification problem) and how these chains interact when triggered by several actions (the concurrent action problem). Elaborating on the works of Dean and Wellman [1991], this paper shows how the frame, ramification, and concurrency problems can be handled effectively using the language of causal graphs, (see also [Darwiche & Pearl 1994]).

The key idea is that causal knowledge can efficiently be organized in terms of just a few basic mechanisms, each involving a relatively small number of variables and each encoded as a set of functional constraints perturbed by random disturbances. Each external elementary action overrules just one mechanism while leaving the others unaltered. The specification of an action then requires only the identification of the mechanisms that are overruled by that action. Once these mechanisms are identified, the effect of the action (or combinations thereof) can be computed from the constraints imposed by the remaining mechanisms.

The semantics behind causal graphs and their relations to actions and belief networks have been discussed in prior publications [Pearl & Verma 1991, Goldszmidt & Pearl 1992, Druzdzel & Simon 1993, Pearl 1993a, Spirtes et al. 1993, Pearl 1993b]. In Spirtes et al. [1993] and later in Pearl [1993b], for example, it was shown how graphical representation can be used to facilitate quantitative predictions of the effects of interventions, including interventions that were not contemplated during the network's construction. Section 2 reviews this aspect of causal networks, following the formulation in [Pearl 1993b].

The main problem addressed in this paper is quantification of the effects of interventions when the causal graph is not fully parameterized, that is, when we are given the topology of the graph but not the conditional probabilities on all variables. In this situation, numerical probabilities are given to only a subset of variables, in the form of unstructured conditional probability sentences. This is a unless you have a comparative realistic setting in AI applications, where the user/designer might not have either the patience or the knowledge necessary for specification of a complete distribution function; some combinations of variables may be too esoteric to be assigned probabilities, and some variables may be too hypothetical (e.g., "life style" or "attitude") to even be parameterized numerically.

---

[2] Causal information can in fact be viewed as an implicit encoding of responses to future actions, and, in practice, causal information is assumed and used by most decision analysts. The ID literature's insistence on divorcing the links in the ID from any causal interpretation [Howard & Matheson 1981, Howard 1989] is, therefore, at odds with prevailing practice. Section 2 of this paper can be viewed as a way to formalize and reinstate the causal reading of influence diagrams.

[3] This assumption corresponds to Dalal's [1988] database update, which uses the Hamming distance to define the "closest world" in Lewis's imaging.

To manage this problem, this paper introduces a calculus that operates on whatever probabilistic and causal information is available and, using symbolic transformations on the input sentences, produces probabilistic assessments of the effects of actions. The calculus admits two types of conditioning operators: ordinary Bayes conditioning, $P(y|X = x)$; and causal conditioning, $P(y|do(X = x))$, that is, the probability of $Y = y$ conditioned on holding $X$ constant (at $x$) by deliberate external action.[4] Given a causal graph and an input set of conditional probabilities, the calculus derives new conditional probabilities of both the Bayesian and the causal types and, whenever possible, generates closed form expressions for the effect of interventions in terms of the input information.

## 2   THE MANIPULATIVE READING OF CAUSAL NETWORKS: A REVIEW

The connection between the probabilistic and the manipulative readings of directed acyclic graphs (DAGs) is formed through Simon's [1977] mechanism-based model of causal ordering.[5] In this model, each child-parent family in a DAG $G$ represents a deterministic function

$$X_i = f_i(\mathbf{pa}_i, \epsilon_i), \qquad (1)$$

where $\mathbf{pa}_i$ are the parents of variable $X_i$ in $G$, and $\epsilon_i$, $0 < i < n$, are mutually independent, arbitrarily distributed random disturbances. A *causal theory* is a pair $< P, G >$, where $G$ is a DAG and $P$ is the probability distribution that results from the functions $f_i$ in (1).

Characterizing each child-parent relationship as a deterministic function, instead of the usual conditional probability $P(x_i \mid \mathbf{pa}_i)$, imposes equivalent independence constraints on the resulting distributions and

---

[4]The notation $set(X = x)$ was used in [Pearl 1993b], while $do(X = x)$ was used in [Goldszmidt and Pearl 1992].

[5]This mechanism-based model was adopted in [Pearl & Verma 1991] for defining probabilistic causal theories. It has been elaborated in Druzdzel & Simon [1993] and is also the basis for the "invariance" principle of Spirtes et al. [1993].

leads to the same recursive decomposition

$$P(x_1, ..., x_n) = \prod_i P(x_i \mid \mathbf{pa}_i) \qquad (2)$$

that characterizes Bayesian networks [Pearl 1988]. This is so because each $\epsilon_i$ is independent on all non-descendants of $X_i$. However, the functional characterization $X_i = f_i(\mathbf{pa}_i, \epsilon_i)$ also specifies how the resulting distribution would change in response to external interventions, since, by convention, each function is presumed to remain constant unless specifically altered. Moreover, the nonlinear character of $f_i$ permits us to treat changes in the function $f_i$ itself as a variable, $F_i$, by writing

$$X_i = f'_i(\mathbf{pa}_i, F_i, \epsilon_i) \qquad (3)$$

where

$$f'_i(a, b, c) = f_i(a, c) \text{ whenever } b = f_i$$

Thus, any external intervention $F_i$ that alters $f_i$ can be represented graphically as an added parent node of $X_i$, and the effect of such an intervention can be analyzed by Bayesian conditionalization, that is, by simply setting this added parent variable to the appropriate value $f_i$.

The simplest type of external intervention is one in which a single variable, say $X_i$, is forced to take on some fixed value, say, $x'_i$. Such intervention, which we call *atomic*, amounts to replacing the old functional mechanism $X_i = f_i(\mathbf{pa}_i, \epsilon_i)$ with a new mechanism $X_i = x'_i$ governed by some external force $F_i$ that sets the value $x'_i$. If we imagine that each variable $X_i$ could potentially be subject to the influence of such an external force $F_i$, then we can view the causal network $G$ as an efficient code for predicting the effects of atomic interventions and of various combinations of such interventions.

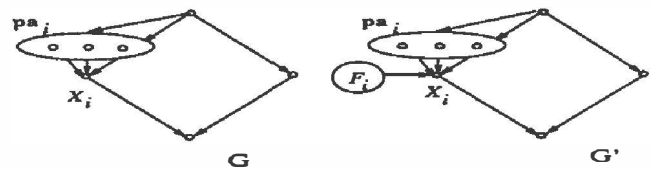

Figure 1: Representing external intervention $F_i$ by an augmented network $G' = G \cup \{F_i \rightarrow X_i\}$.

The effect of an atomic intervention $do(X_i = x'_i)$ is encoded by adding to $G$ a link $F_i \longrightarrow X_i$ (see Figure 1), where $F_i$ is a new variable taking values in $\{do(x'_i), idle\}$, $x'_i$ ranges over the domain of $X_i$, and *idle* represents no intervention. Thus, the new parent set of $X_i$ in the augmented network is $\mathbf{pa}'_i = \mathbf{pa}_i \cup \{F_i\}$, and it is related to $X_i$ by the conditional probability

$$P(x_i \mid \mathbf{pa}'_i)$$
$$= \begin{cases} P(x_i \mid \mathbf{pa}_i) & \text{if } F_i = idle \\ 0 & \text{if } F_i = do(x'_i) \text{ and } x_i \neq x'_i \\ 1 & \text{if } F_i = do(x'_i) \text{ and } x_i = x'_i \end{cases} \qquad (4)$$

The effect of the intervention $do(x_i')$ is to transform the original probability function $P(x_1, ..., x_n)$ into a new function $P_{x_i'}(x_1, ..., x_n)$, given by

$$P_{x_i'}(x_1, ..., x_n) = P'(x_1, ..., x_n \mid F_i = do(x_i')) \quad (5)$$

where $P'$ is the distribution specified by the augmented network $G' = G \cup \{F_i \to X_i\}$ and Eq. (4), with an arbitrary prior distribution on $F_i$. In general, by adding a hypothetical intervention link $F_i \to X_i$ to each node in $G$, we can construct an augmented probability function $P'(x_1, ..., x_n; F_1, ..., F_n)$ that contains information about richer types of interventions. Multiple interventions would be represented by conditioning $P'$ on a subset of the $F_i$'s (taking values in their respective $do(x_i')$), while the pre-intervention probability function $P$ would be viewed as the posterior distribution induced by conditioning each $F_i$ in $P'$ on the value *idle*.

This representation yields a simple and direct transformation between the pre-intervention and the post-intervention distributions:[6]

$$P_{x_i'}(x_1, ..., x_n) = \begin{cases} \frac{P(x_1,...,x_n)}{P(x_i \mid \mathbf{pa}_i)} & \text{if } x_i = x_i' \\ 0 & \text{if } x_i \neq x_i' \end{cases} \quad (6)$$

This transformation reflects the removal of the term $P(x_i \mid \mathbf{pa}_i)$ from the product decomposition of Eq. (2), since $\mathbf{pa}_i$ no longer influence $X_i$. Graphically, the removal of this term is equivalent to removing the links between $\mathbf{pa}_i$ and $X_i$, while keeping the rest of the network intact.

The transformation (6) exhibits the following properties:

1. An intervention $do(x_i)$ can affect only the descendants of $X_i$ in $G$.

2. For any set $\mathbf{S}$ of variables, we have

$$P_{x_i}(\mathbf{S} \mid \mathbf{pa}_i) = P(\mathbf{S} \mid x_i, \mathbf{pa}_i) \quad (7)$$

In other words, given $X_i = x_i$ and $\mathbf{pa}_i$, it is superfluous to find out whether $X_i = x_i$ was established by external intervention or not. This can be seen directly from the augmented network $G'$ (see Figure 1), since $\{X_i\} \cup \mathbf{pa}_i$ d-separates $F_i$ from the rest of the network, thus legitimizing the conditional independence $\mathbf{S} \perp\!\!\!\perp F_i \mid (X_i, \mathbf{pa}_i)$.

3. A sufficient condition for an external intervention $do(X_i = x_i)$ to have the same effect on $X_j$ as the passive observation $X_i = x_i$ is that $X_i$ d-separates $\mathbf{pa}_i$ from $X_j$, that is,

$$P'(x_j \mid do(x_i)) = P(x_j \mid x_i) \text{ iff } X_j \perp\!\!\!\perp \mathbf{pa}_i \mid X_i \quad (8)$$

---

[6]Eq. (6) is a special case of the Manipulation Theorem of Spirtes et al. [1993] which deals with interventions that modify several conditional probabilities simultaneously. According to this source, Eq. (6) was "independently conjectured by Fienberg in a seminar in 1991". An additive version of Eq. (6) was independently presented in [Goldszmidt & Pearl 1992].

The immediate implication of Eq. (6) is that, given the structure of the causal network $G$, one can infer post-intervention distributions from pre-intervention distributions; hence, we can reliably estimate the effects of interventions from passive (i.e., nonexperimental) observations. However, use of Eq. (6) is limited for several reasons. First, the formula was derived under the assumption that the pre-intervention probability $P$ is given by the product of Eq. (2), which represents general domain knowledge prior to making any specific observation. Second, the formula in Eq. (6) is not very convenient in practical computations, since the joint distribution $P(x_1, ..., x_n)$ is represented not explicitly but implicitly, in the form of probabilistic sentences from which it can be computed. Finally, the formula in Eq. (6) presumes that we have sufficient information at hand to define a complete joint distribution function. In practice, a complete specification of $P$ is rarely available, and we must predict the effect of actions from a knowledge base containing unstructured collection of probabilistic statements, some observational and some causal.

The first issue is addressed in [Pearl 1993a and Balke & Pearl 1994], where assumptions about persistence are added to the knowledge base to distinguish properties that terminate as a result of an action from those that persist despite that action. This paper addresses the latter two issues It offers a set of sound (and possibly complete) inference rules by which probabilistic sentences involving actions and observations can be transformed to other such sentences, thus providing a syntactic method of deriving (or verifying) claims about actions and observations. We will assume, however, that the knowledge base contains the topological structure of the causal network $G$, that is, some of its links are annotated with conditional probabilities while others remain unspecified. Given such a partially specified causal theory, our main problem will be to facilitate the syntactic derivation of expressions of the form $P(x_j \mid do(x_i))$.

## 3 A CALCULUS OF ACTIONS

### 3.1 PRELIMINARY NOTATION

Let $X, Y, Z, W$ be four arbitrary disjoint sets of nodes in the DAG $G$. We say that $X$ and $Y$ are independent given $Z$ in $G$, denoted $(X \perp\!\!\!\perp Y \mid Z)_G$, if the set $Z$ d-separates $X$ from $Y$ in $G$. We denote by $G_{\overline{X}}$ ($G_{\underline{X}}$, respectively) the graph obtained by deleting from $G$ all arrows pointing to (emerging from, respectively) nodes in $X$.

Finally, we replace the expression $P(y \mid do(x), z)$ by a shorter expression $P(y \mid \hat{x}, z)$, using the $\hat{\ }$ symbol to identify the variables that are kept constant externally. In words, the expression $P(y \mid \hat{x}, z)$ stands for the probability of $Y = y$ given that $Z = z$ is observed and $X$ is held constant at $x$.

458  Pearl

## 3.2 INFERENCE RULES

Armed with this notation, we are now able to formulate the three basic inference rules of the proposed calculus.

**Theorem 3.1** *Given a causal theory $< P, G >$, for any sets of variables $X, Y, Z, W$ we have:*

**Rule 1** *Insertion/deletion of observations (Bayes conditioning)*

$$P(y|\hat{x}, z, w) = P(y|\hat{x}, w) \quad \text{if} \quad (Y \parallel Z|X, W)_{G_{\overline{X}}}$$

**Rule 2** *Action/observation exchange*

$$P(y|\hat{x}, \hat{z}, w) = P(y|\hat{x}, z, w) \quad \text{if} \quad (Y \parallel Z|X, W)_{G_{\overline{X}\underline{Z}}}$$

**Rule 3** *Insertion/deletion of actions*

$$P(y|\hat{x}, \hat{z}, w) = P(y|\hat{x}, w) \quad \text{if} \quad (Y \parallel Z|X, W)_{G_{\overline{X}\,\overline{Z(W)}}}$$

*where $Z(W)$ is the set of $Z$ nodes that are not ancestors of any $W$ node in $G_{\overline{X}}$.*

Each of the inference rules above can be proven from the basic interpretation of the "do($x$)" operation as a replacement of the causal mechanism that connects $X$ to its parent prior to the action with a new mechanism $X = x$ introduced by the intervening force (as in Eqs. (4) - (5)).

Rule 1 reaffirms $d$-separation as a legitimate test for Bayesian conditional independence in the distribution determined by the intervention $do(X = x)$, hence the graph $G_{\overline{X}}$.

Rule 2 provides conditions for an external intervention $do(Z = z)$ to have the same effect on $Y$ as the passive observation $Z = z$. The condition is equivalent to Eq. (8), since $G_{\overline{X}\underline{Z}}$ eliminates all paths from $Z$ to $Y$ (in $G_{\overline{X}}$) which do not go through $pa_Z$.[7]

Rule 3 provides conditions for introducing (or deleting) an external intervention $do(Z = z)$ without affecting the probability of $Y = y$. Such operation would be valid if the $d$-separation $(Y \parallel F_Z|X, W)$ is satisfied in the augmented graph $G'_{\overline{X}}$, since it implies that the manipulating variables $F_Z$ have no effect on $Y$. The condition used in Rule 3, $(Y \parallel Z|X, W)_{G_{\overline{X}\,\overline{Z(W)}}}$, translates the one above into $d$-separation between $Y$ and $Z$ (in the unaugmented graph) by pruning the appropriate links entering $Z$.

---
[7]This condition was named the "back-door" criterion in [Pearl 1993b], echoing the requirement that only indirect paths from $Z$ to $Y$ be $d$-separated; these paths can be viewed as entering $Z$ through the back door. An equivalent, though more complicated, graphical criterion is given in Theorem 7.1 of [Spirtes et al. 1993].

## 3.3 EXAMPLE

We will now demonstrate how these inference rules can be used to quantify the effect of actions, given partially specified causal theories. Consider the causal theory $< P(x, y, z), G >$, where $G$ is the graph given in Figure 2 below and $P(x, y, z)$ is the distribution over the observed variables $X, Y, Z$. Since $U$ is unobserved, the theory is only partially specified; it will be impossible to infer all required parameters, such as $P(u)$ or $P(y|z, u)$. We will see, however, that this structure still permits us to quantify, using our calculus, the effect of every action on every observed variable.

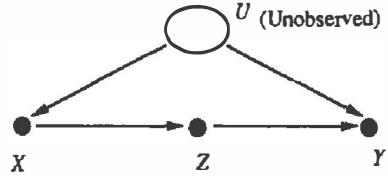

Figure 2

The applicability of each inference rule requires that certain $d$-separation conditions hold in some graph, whose structure will vary with the expressions to be manipulated. Figure 3 displays the graphs that will be needed for the derivations that follow.

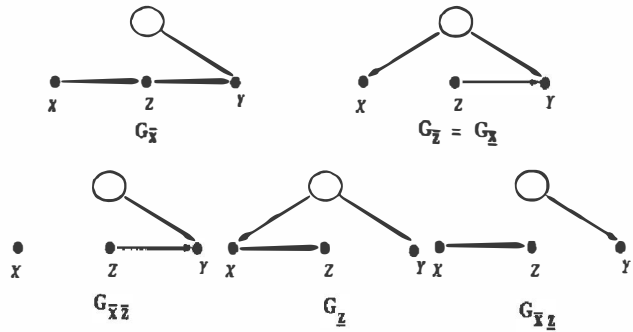

Figure 3

**Task-1, compute $P(z|\hat{x})$**
This task can be accomplished in one step, since $G$ satisfies the applicability condition for Rule 2, namely $X \parallel Z$ in $G_{\underline{X}}$ (because the path $X \leftarrow U \rightarrow Y \leftarrow Z$ is blocked by the collider at $Y$), and we can write

$$P(z|\hat{x}) = P(z|x) \qquad (9)$$

**Task-2, compute $P(y|\hat{z})$**
Here we cannot apply Rule 2 to substitute $\hat{z}$ for $z$ because $G_{\underline{Z}}$ contains a back-door path from $Z$ to $Y$. Naturally, we would like to "block" this path by conditioning on variables (such as $X$) that reside on that path. Symbolically, this operation involves conditioning and summing over all values of $X$,

$$P(y|\hat{z}) = \sum_x P(y|x, \hat{z})P(x|\hat{z}) \qquad (10)$$

We now have to deal with two expressions involving $\hat{z}$, $P(y|x,\hat{z})$ and $P(x|\hat{z})$. The latter can be readily computed by applying Rule 3 for action deletion:

$$P(x|\hat{z}) = P(x) \text{ if } (Z \perp\!\!\!\perp X)_{G_{\overline{Z}}} \qquad (11)$$

noting that, indeed, $X$ and $Z$ are $d$-separated in $G_{\overline{Z}}$. (This can be seen immediately from Figure 2; manipulating $Z$ will have no effect on $X$.) To reduce the former quantity, $P(y|x,\hat{z})$, we consult Rule 2

$$P(y|x,\hat{z}) = P(y|x,z) \text{ if } (Z \perp\!\!\!\perp Y|X)_{G_{\underline{Z}}} \qquad (12)$$

and note that $X$ $d$-separates $Z$ from $Y$ in $G_{\underline{Z}}$. This allows us to write Eq. (10) as

$$P(y|\hat{z}) = \sum_x P(y|x,z)P(x) = E_x P(y|x,z) \qquad (13)$$

which is a special case of the back-door formula [Pearl 1993b, Eq. (11)] with $S = X$. This formula appears in a number of treatments on causal effects (e.g., [Rosenbaum & Rubin 1983, Pratt & Schlaifer 1988, Rosenbaum 1989;]) where the legitimizing condition, $(Z \perp\!\!\!\perp Y|X)_{G_{\underline{Z}}}$, was given a variety of names, all based on conditional-independence judgments about counterfactual variables. Action calculus replaces such judgments by formal tests ($d$-separation) on a single graph ($G$) that represents the domain knowledge.

We are now ready to tackle the evaluation of $P(y|\hat{x})$, which cannot be reduced to an observational expression by direct application of any of the inference rules.

**Task-3, compute $P(y|\hat{x})$**

Writing

$$P(y|\hat{x}) = \sum_z P(y|z,\hat{x})P(z|\hat{x}) \qquad (14)$$

we see that the term $P(z|\hat{x})$ was reduced in Eq. (9) while no rule can be applied to eliminate the manipulation symbol $\hat{\ }$ from the term $P(y|z,\hat{x})$. However, we can add a $\hat{\ }$ symbol to this term via Rule 2

$$P(y|z,\hat{x}) = P(y|\hat{z},\hat{x}) \qquad (15)$$

since Figure 3 shows

$$(Y \perp\!\!\!\perp Z|X)_{G_{\overline{XZ}}}$$

We can now delete the action $\hat{x}$ from $P(y|\hat{z},\hat{x})$ using Rule 3, since $Y \perp\!\!\!\perp X|Z$ holds in $G_{\overline{XZ}}$. Thus, we have

$$P(y|z,\hat{x}) = P(y|\hat{z}) \qquad (16)$$

which was calculated in Eq. (13). Substituting Eqs. (13), (16), and (9) back into Eq. (14) finally yields

$$P(y|\hat{x}) = \sum_z P(z|x) \sum_{x'} P(y|x',z)P(x') \qquad (17)$$

In contrast to the back-door formula of Eq. (13), Eq. (17) computes the causal effect of $X$ on $Y$ using an intermediate variable $Z$ that is affected by $X$.

**Task-4, compute $P(y,z|\hat{x})$**

$$P(y,z|\hat{x}) = P(y|z,\hat{x})P(z|\hat{x}) \qquad (18)$$

The two terms on the r.h.s. were derived before in Eqs. (9) and (16), from which we obtain

$$\begin{aligned} P(y,z|\hat{x}) &= P(y|\hat{z})P(z|x) \\ &= P(z|x) \sum_{x'} P(y|x',z)P(x') \end{aligned}$$

### 3.4 DISCUSSION

Computing the effects of actions by using partial theories in which probabilities are specified on a select subset of (observed) variables is an extremely important task in statistics and socio-economic modeling, since it determines when causal effects are "identifiable" (i.e., estimable consistently from non-experimental data) and this when randomized experiments are not needed. The calculus proposed here, reduces the problem of identifiability to the problem of finding a sequence of transformations, each conforming to one of the inference rules in Theorem 3.1, which reduces an expression of the form $P(y|\hat{x})$ to a standard (i.e., hat-free) probability expression. Note that whenever a reduction is possible, the calculus provides a closed form expression for the desired causal effect.

The proposed calculus uncovers many new structures that permit the identification of causal effects from nonexperimental observations. For example, the structure of Figure 3 represents a large class of observational studies in which the causal effect of an action ($X$) can be determined by measuring a variable ($Z$) that mediates the interaction between the action and its effect ($Y$). Most of the literature on statistical experimentation considers the measurement of intermediate variables, affected by the action, to be useless, if not harmful, for causal inference [Cox 1958, Pratt & Schlaifer 1988]. The relevance of such structures in practical situations can be seen, for instance, if we identify $X$ with smoking, $Y$ with lung cancer, $Z$ with the amount of tar deposited in a subject's lungs, and $U$ with an unobserved carcinogenic genotype that, according to the tobacco industry, also induces an inborn craving for nicotine. In this case, Eq. (17) would provide us with the means to quantify, from nonexperimental data, the causal effect of smoking on cancer. (Assuming, of course, that the data $P(x,y,z)$ is made available and that we believe that smoking does not have any direct causal effect on lung cancer except that mediated by tar deposits).

In this example, we were able to compute answers to all possible queries of the form $P(y|z,\hat{x})$ where $Y$, $Z$, and $X$ are subsets of observed variables. In general, this will not be the case. For example, there is no general way of computing $P(y|\hat{x})$ from the observed distribution whenever the causal model contains the subgraph shown in Figure 4, where $X$ and $Y$ are adja-cent and the dashed line represents a path traversing unobserved variable.[8] Similarly, our ability to compute $P(y|\hat{x})$ for every pair of singleton variables does not ensure our ability to compute joint distributions, such as $P(y_1, y_2|\hat{x})$. Figure 5, for example, shows a causal graph where both $P(y_1|\hat{x})$ and $P(y_2|\hat{x})$ are computable, but $P(y_1, y_2|\hat{x})$ is not; consequently, we cannot compute $P(z|\hat{x})$. Interestingly, the graph of Figure 5 is the smallest graph that does not contain the pattern of Figure 4 and still presents an uncomputable causal effect. Graphical criteria for identifiability and nonidentifiability are given in [Pearl 1994].

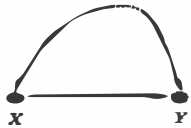 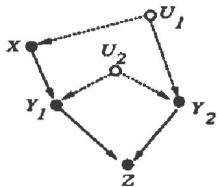

Figure 4    Figure 5

Another interesting feature demonstrated by the network in Figure 5 is that it is often easier to compute the effects of a joint action than the effects of its constituent singleton actions[9]. In this example, it is possible to compute $P(z|\hat{x}, \hat{y}_2)$ and $P(z|\hat{x}, \hat{y}_1)$, yet there is no way of computing $P(z|\hat{x})$. For example, the former can be evaluated by invoking Rule 2, giving

$$P(z|\hat{x}, \hat{y}_2) = \sum_{y_1} P(z|y_1, \hat{x}, \hat{y}_2) P(y_1|\hat{x}, \hat{y}_2)$$
$$= \sum_{y_1} P(z|y_1, x_1, y_2) P(y_1|x)$$

On the other hand, Rule 2 cannot be applied to the computation of $P(y_1|\hat{x}, y_2)$ because, conditioned on $Y_2$, $X$ and $Y_1$ are d-connected in $G_{\underline{X}}$ (through the dashed lines). We conjecture, however, that whenever $P(y|\hat{x}_i)$ is computable for every singleton $x_i$, then $P(y|\hat{x}_1, \hat{x}_2, ..., \hat{x}_l)$ is computable as well, for any subset of variables $\{X_1, ..., X_l\}$.

Our calculus is not limited to the derivation of causal probabilities from noncausal probabilities; we can derive conditional and causal probabilities from causal expressions as well. For example, given the graph of Figure 2 together with the quantities $P(z|\hat{x})$ and $P(y|\hat{z})$, we can derive an expression for $P(y|\hat{x})$,

$$P(y|\hat{x}) = \sum_z P(y|\hat{z}) P(z|\hat{x}) \qquad (19)$$

---

[8] One can calculate strict upper and lower bounds on $P(y|\hat{x})$ and these bounds may coincide for special distributions, $P(x, y, z)$ [Balke & Pearl 1994], but there is no way of computing $P(y|\hat{x})$ for *every* distribution $P(x, y, z)$.

[9] The fact that the two tasks are not equivalent was brought to my attention by James Robins, who has worked out many of these computations in the context of sequential treatment management [Robins 1989].

using the steps that led to Eq. (16). Note that this derivation is still valid when we add a common cause to $X$ and $Z$, which is the most general condition under which the transitivity of causal relationships holds. In [Pearl 1994] we present conditions for transforming $P(y|\hat{x})$ into expressions in which only members of $Z$ obtain the hat symbol. These would enable an agent to measure $P(y|\hat{x})$ by manipulating a surrogate variable, $Z$, which is easier to control than $X$.

### 3.5 CONDITIONAL ACTIONS AND STOCHASTIC POLICIES

The interventions considered thus far were unconditional actions that merely force a variable or a group of variables $X$ to take on some specified value $x$. In general, interventions may involve complex policies in which a variable $X$ is made to respond in a specified way to some set $Z$ of other variables, say through a functional relationship $X = g(Z)$ or through a stochastic relationship whereby $X$ is set to $x$ with probability $P^*(x|z)$. We will show that computing the effect of such policies is equivalent to computing the expression $P(y|\hat{x}, z)$.

Let $P(y|do(X = g(Z)))$ stand for the distribution (of $Y$) prevailing under the policy $(X = g(Z))$. To compute $P(y|do(X = g(Z)))$, we condition on $Z$ and write

$$\begin{aligned} &P(y|do(X = g(Z))) \\ &= \sum_z P(y|do(X = g(z)), z) P(z|do(X = g(z))) \\ &= \sum_z P(y|\hat{x}, z)|_{x=g(z)} P(z) \\ &= E_z[P(y|\hat{x}, z)|_{x=g(z)}] \end{aligned}$$

The equality

$$P(z|do(X = g(z))) = P(z)$$

stems, of course, from the fact that $Z$ cannot be a descendant of $X$, hence, whatever control one exerts on $X$, it can have no effect on the distribution of $Z$.

Thus, we see that the causal effect of a policy $X = g(Z)$ can be evaluated directly from the expression of $P(y|\hat{x}, z)$, simply by substituting $g(z)$ for $x$ and taking the expectation over $Z$ (using the observed distribution $P(z)$).

The identifiability condition for policy intervention is somewhat stricter than that for a simple intervention. Clearly, whenever a policy $do(X = g(Z))$ is identifiable, the simple intervention $do(X = x)$ is identifiable as well, as we can always get the latter by setting $g(Z) = X$. The converse, does not hold, however, because conditioning on $Z$ might create dependencies that will prevent the successful reduction of $P(y|\hat{x}, z)$ to a hat-free expression.

A stochastic policy, which imposes a new conditional distribution $P^*(x|z)$ for $x$, can be handled in a similar manner. We regard the stochastic intervention as a random process in which the unconditional intervention $do(X = x)$ is enforced with probability $P^*(x|z)$. Thus, given $Z = z$, the intervention $set(X = x)$ will occur with probability $P^*(x|z)$ and will produce a causal effect given by $P(y|\hat{x}, z)$. Averaging over $x$ and $z$ gives

$$P(y|P^*(x|z)) = \sum_x \sum_z P(y|\hat{x}, z) P^*(x|z) P(z)$$

Since $P^*(x|z)$ is specified externally, we see again that the identifiability of $P(y|\hat{x}, z)$ is a necessary and sufficient condition for the identifiability of any stochastic policy that shapes the distribution of X by the outcome of $Z$.

Of special importance in planning is a STRIP-like action whose immediate effects $X = x$ depend on the satisfaction of some enabling precondition $C(w)$ on a set $W$ of variables. To represent such actions, we let $Z = W \cup \text{pa}_X$ and set

$$P^*(x|z) = \begin{cases} P(x|\text{pa}_X) & \text{if } C(w) = false \\ 1 & \text{if } C(w) = true \text{ and } X = x \\ 0 & \text{if } C(w) = true \text{ and } X \neq x \end{cases}$$

It should be noted, however, that in planning applications the effect of an action may be to invalidate its preconditions. To represent such actions, temporally indexed causal networks are necessary [Dean & Kanazawa 1989, Pearl 1993a, Balke & Pearl 1994].

## 4 CONCLUSIONS

The calculus proposed in this paper captures in symbols and graphs the conceptual distinction between *seeing* and *doing*. While many systems have implemented this obvious distinction—from early systems of adaptive control to their modern AI counterparts of [Dean and Kanazawa 1989] and [Draper et al. 1994]— the belief-changing operators of seeing and doing can now enjoy the power of symbolic manipulations. The calculus permits the derivation of expressions for states of belief that result from sequences of actions and observations, which, in turn, should permit the identification of variables and relationships that are crucial for the success of a given plan or strategy. The exercise in Section 3.3, for example, demonstrates how predictions about the effects of actions can be derived from passive observations even though portions of the knowledge base (connected with the unobserved variable U) remain inaccessible. Another possible application of the proposed calculus lies in the area of learning, where it might facilitate the integration of the two basic modes of human learning: learning by manipulation and learning by observation.

The immediate beneficiaries of the proposed calculus would be social scientists and clinical trilists, as the calculus enables experimental researchers to translate complex considerations of causal interactions into a formal language, thus facilitating the following tasks:

1. Explicate the assumptions underlying the model.
2. Decide whether the assumptions are sufficient for obtaining consistent estimates of the target quantity: the total effect of one variable on another.
3. If the answer to item 2 is affirmative, the method provides a closed-form expression for the target quantity, in terms of distributions of observed quantities.
4. If the answer to item 2 is negative, the method suggests a set of observations and experiments which, if performed, would render a consistent estimate feasible.

The bizzare confusion and controversy surrounding the role of causality in statistics stems largely from the lack of mathematical notation for defining, expressing, and manipulating causal relationships. Statisticians will benefit, therefore, from a calculus that integrates both statistical and causal information, and in which causal influences are kept distinct from probabilistic dependencies.

There are also direct applications of action calculus to expert systems and Bayesian networks technology. One conceptual contribution, mentioned in Section 1, is the appeal to causality for inferring the effect of certain actions without those actions being explicitly encoded in the knowledge base. This facility simplifies the knowledge elicitation process by focusing attention on causal relationships and by dispensing with the specification of actions whose effects can be inferred from those relationships.

A second contribution involves the treatment of hidden variables. Such variables represent factors that the expert chooses to exclude from formal analysis, either because they lie beyond the scope of the domain or because they are inaccessible to measurement. The example of Section 3.3 demonstrates that certain queries can be answered precisely without the parameters associated with hidden variables assessing. Action calculus should identify the conditions under which such assessments can be saved.

### Acknowledgments

This investigation was inspired by Goldszmidt's formalization of actions in nonmonotonic reasoning [Goldszmidt 1992] and by [Spirtes et al. 1993], in which a graphical account of manipulations was first proposed. The investigation also benefitted from discussions with Adnan Darwiche, Phil Dawid, Arthur Goldberger, Ed Leamer, John Pratt, James Robins, Donald Rubin, Keunkwan Ryu, Glenn Shafer, and Michael Sobel. The research was partially supported by Air Force grant #F49620-94-1-0173, NSF grant #IRI-9200918, and Northrop-Rockwell Micro grant #93-124.